\pdfoutput=1

\documentclass[11pt]{article}

\usepackage[final]{acl}

\usepackage{times}
\usepackage{latexsym}

\usepackage[T1]{fontenc}

\usepackage[utf8]{inputenc}

\usepackage{microtype}

\usepackage{inconsolata}

\usepackage{graphicx}

\usepackage{mdframed}
\usepackage{arydshln}
\usepackage{enumitem}
\setlist[enumerate]{itemsep=0pt, topsep=0pt, parsep=0pt}
\usepackage{xcolor}

\title{Chain-of-Description: What I can understand, I can put into words}

\author{Jiaxin GUO, Daimeng Wei, Zongyao Li, Hengchao Shang, Yuanchang Luo, Hao Yang\\
        \{jiaxinguo1,weidaimeng,lizongyao,shanghengchao,luoyuanchang1,yanghao30\}@huawei.com\\
        Huawei Translation Services Center, Beijing, China }

\begin{document}
\maketitle
\begin{abstract}

In this paper, we propose a novel strategy defined as Chain-of-Description (CoD) Prompting, tailored for Multi-Modal Large Language Models. This approach involves having the model first provide a detailed description of the multi-modal input before generating an answer to the question. When applied to models such as Qwen2-Audio, Qwen2-VL, and Qwen2.5-VL, CoD Prompting significantly enhances performance compared to standard prompting methods. This is demonstrated by nearly a 4\% improvement in the speech category of the audio benchmark AIR-Bench-Chat and a 5.3\% improvement in the hard-level portion of the vision benchmark MMMU\_Pro. Our ablation study further validates the effectiveness of CoD Prompting.
\end{abstract}

\section{Introduction}


Multi-Modal Large Language Models \cite{DBLP:conf/icml/Wu0Q0C24,DBLP:journals/corr/abs-2306-13549,DBLP:journals/corr/abs-2409-18042} (MLLMs), which encompass Large Audio-Language Models \cite{DBLP:journals/corr/abs-2408-16725,DBLP:journals/corr/abs-2405-08295,DBLP:journals/corr/abs-2409-06666,DBLP:conf/asru/WuGCZZWLLRLW23,DBLP:journals/corr/abs-2311-07919,DBLP:journals/corr/abs-2407-10759} (LALMs) and Large Vision-Language Models \cite{DBLP:journals/corr/abs-2312-14238,DBLP:journals/corr/abs-2404-16821,DBLP:journals/corr/abs-2408-03326,DBLP:journals/corr/abs-2412-15188,DBLP:journals/corr/abs-2409-12191,Qwen2.5-VL} (LVLMs), have shown considerable potential in managing a variety of input types. These models are generally based on Large Language Models (LLMs) and employ audio/vision encoders to align multi-modal inputs with text. However, the conventional method of directly generating answers from audio or visual inputs might not fully capitalize on the information and comprehension that the models can potentially extract. Given that the training paradigm for most MLLMs involves aligning multi-modal inputs with text, we explored whether there is an inference strategy that can not only explicitly align these inputs but also improve the quality of the results produced by MLLMs.


We believe that \textit{"What I can understand, I can put into words."} This implies that if a model can generate a detailed description of the input, it indicates a deeper level of understanding. Based on this idea, we proposed the Chain-of-Description (CoD) prompting for MLLMs, which involves having the model first provide a detailed description of the multi-modal input before answering the question. 



We have implemented CoD Prompting on LALMs and LVLMs, specifically with the Qwen2-Audio \cite{DBLP:journals/corr/abs-2407-10759}, Qwen2-VL \cite{DBLP:journals/corr/abs-2409-12191}, and Qwen2.5-VL \cite{Qwen2.5-VL} models. Through experiments conducted on their respective test sets, we have demonstrated that CoD significantly enhances the reasoning performance of these models compared to the standard approach. Specifically, in the speech testset AIR-Bench-Chat \cite{DBLP:conf/acl/YangXLC0ZLLZZZ24}, the Qwen2-Audio model performance for human speech improved by nearly 4\%. In the image testset MMLU\_Pro \cite{DBLP:journals/corr/abs-2409-02813}, for the hard-level category, the performance of Qwen*-VL increased by 5.3\%. 


Furthermore, in our experiments with LALMs, we analyzed information density to substantiate why CoD is effective. In our experiments with LVLMs, we verified the correctness of CoD by constructing better descriptions.

\section{Chain-of-Description Prompting}

\begin{figure*}[t]
    \centering
    \includegraphics[width=15cm]{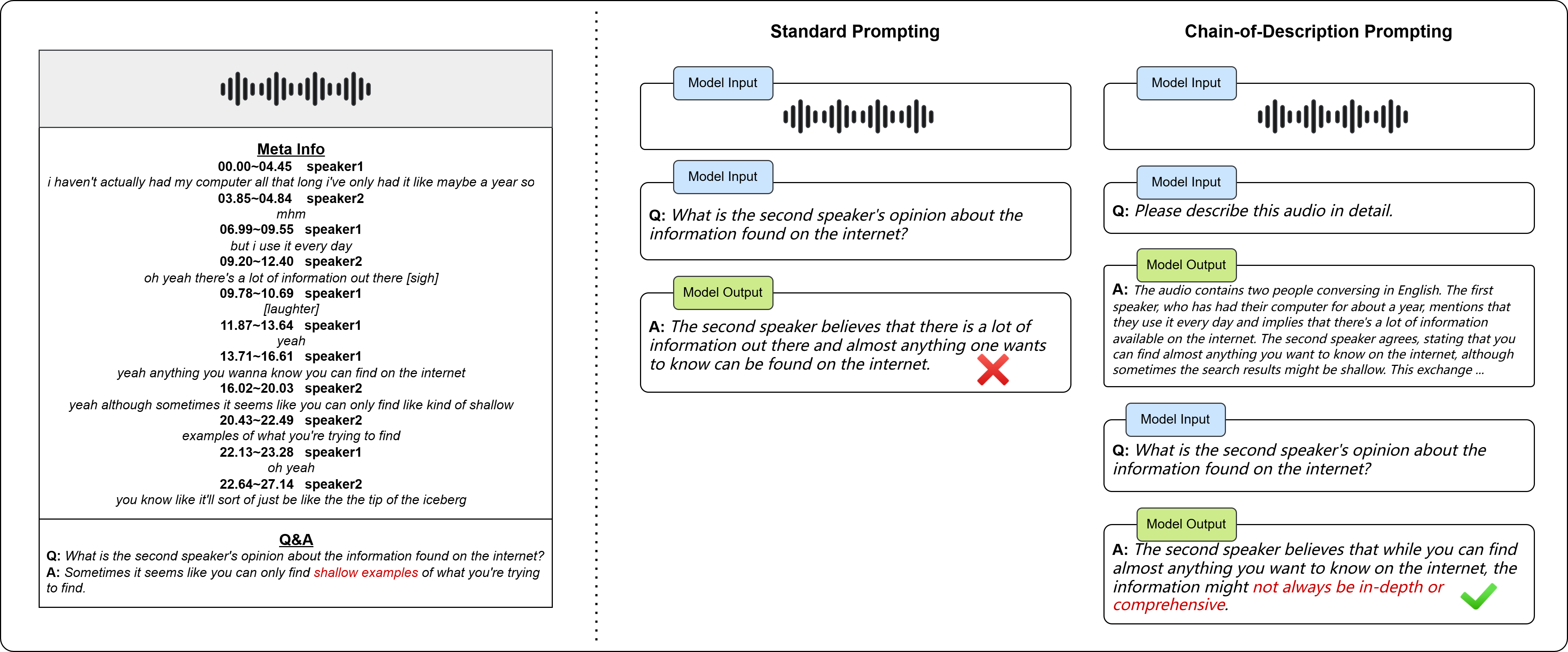}
    \caption{An example of using Standard Prompting and our Chain-of-Description (CoD) Prompting for Large Audio-Language Models (LALMs).}
    \label{fig:CoD_LALM}
\end{figure*}

\begin{mdframed}[backgroundcolor=gray!10]
    \textbf{Motivation:} \textit{What I can understand, I can put into words.}
\end{mdframed}

The motivation of our proposed Chain-of-Description (CoD) Prompting is if a model can generate a detailed description of the input, it indicates a deeper level of understanding. CoD Prompting can be described as follows:
\begin{enumerate}
    \item Supply audio/vision inputs to MLLLMs and ask the models produce detailed descriptions.
    \item MLLMs generate a comprehensive textual representation.
    \item Thereafter, introduce the queries into MLLMs.
    \item MLLMs generate responses pertinent to the queries.
\end{enumerate}

For LALMs, describing speech context, background sounds, and other audio features comprehensively helps the model better understand audio inputs. Figure \ref{fig:CoD_LALM} illustrates examples of using Standard Prompting and CoD Prompting for LALMs. Likewise, for LVLMs, refer to Figure \ref{fig:CoD_LVLM} in Appendix \ref{sec:appendix1}, detailing objects, scenes, colors, and spatial relationships in images enhances comprehension. Focusing on the description process first aims to establish a strong foundation for generating higher-quality answers, improving MLLMs' overall performance.

\section{Experiments with CoD in LALMs}


\subsection{Experimental Setup}

\paragraph{Model}
The model used for our experiments is Qwen2-Audio\footnote{https://huggingface.co/Qwen/Qwen2-Audio-7B-Instruct}, a state-of-the-art open-sourced LALM capable of processing various audio inputs and generating textual responses. 

\paragraph{Evaluation Dataset}

The dataset we utilized is the AIR-Bench\footnote{https://github.com/OFA-Sys/AIR-Bench}, which is the first and widely adopted benchmark designed to assess the comprehension capabilities of LALMs across various audio signals, including human speech, natural sounds, and music. Following prior research, we conducted detailed evaluations on all four subcategories of the AIR-Bench Chat Benchmark (AIR-Bench-Chat): Speech, Sound, Music, and Mixed Audio. More see Appendix \ref{sec:appendix_case_lalms}.

\paragraph{Evaluation Method}

Building upon previous work \cite{DBLP:journals/corr/abs-2407-10759,DBLP:conf/acl/YangXLC0ZLLZZZ24}, we employed an evaluation method that utilizes a LLM as the judge \cite{DBLP:journals/corr/abs-2412-05579,DBLP:journals/corr/abs-2411-15594}. Specifically, we utilized a LLM to rate both the ground truth answer and the model prediction on a scale of 1 to 10. The final score is the average of these ratings. Considering cost-effectiveness, we chose gpt-4o-mini as our evaluation LLM.

In practice, we rated the model predictions from both Standard Prompting and CoD Prompting against the ground truth answer. This process yielded two sets of scores for the ground truth answer, which may exhibit minor differences. To facilitate more effective comparison, we calculated the ratio $r$ of the model prediction $p$ score $s_{p}$ to the ground truth answer $gt$ score $s_{gt}$. The $r$ quantifies the alignment between the prediction and the ground truth. See Appendix \ref{sec:appendix2}.




\subsection{Results}

\begin{table}[h]
    \centering
    \begin{tabular}{ccccc}
    \hline
     & Speech & Sound & Music & Mixed \\
    \hline
    \multicolumn{5}{c}{\textit{Standard Prompting}} \\
    \hdashline
    $s_{gt}$ & 8.23 & 8.01 & 7.96 & 8.38 \\
    $s_{p}$ & 7.51 & 6.93 & 6.87 & 6.49 \\
    $r$ & 91.24\% & 86.48\% & 86.31\% & 77.41\% \\
    \hline
    \multicolumn{5}{c}{\textit{CoD Prompting}} \\
    \hdashline
    $s_{gt}$ & 8.04 & 7.77 & 8.12 & 8.12 \\
    $s_{p}$ & 7.64 & 6.83 & 7.08 & 6.38 \\
    $r$ & \textbf{95.02\%} & \textbf{87.87\%} & \textbf{87.22\%} & \textbf{78.50\%} \\
    \hline
    \end{tabular}
    \caption{The evaluation results for LALMs. These results are based on the AIR-Bench-Chat dataset among four subcategories: Speech, Sound, Music, and Mixed, where the \textit{gpt-4o-mini} rated the model predictions ($p$) of the \textit{Qwen2-Audio} model under both Standard Prompting and CoD Prompting and ground truth answers ($gt$) on a scale of 1 to 10. The average scores for the ground truth answer and model prediction are denoted as $s_{gt}$ and $s_{p}$. The alignment between the prediction and the ground truth is measured by $r=\frac{s_{p}}{s_{gt}}$.}
    \label{tab:rst_lalm_1}
\end{table}

Based on the evaluation results presented in Table \ref{tab:rst_lalm_1}, it is evident that the Qwen2-Audio model's alignment with the ground truth answers has improved across all subcategories after adopting the CoD Prompting. Particularly in the Speech category, the alignment reached 95.02\%, marking an increase of nearly 4\% compared to the Standard Prompting. Other categories such as Sound, Music, and Mixed also experienced an enhancement of about 1\% each. On average across all categories, there was an increase of 1.79\%. This indicates that the CoD Prompting method significantly enhances the consistency of the model's predictions with the ground truth answers. More in Appendix \ref{sec:appendix3}.

\subsection{Ablation Study}

\begin{mdframed}[backgroundcolor=gray!10]
    \textbf{Question:} \textit{Why does CoD Prompting perform better in the Speech category?}
\end{mdframed}




We analyze information density to explain why CoD Prompting performs better in the Speech category. We can consider the description as a textual representation of the audio, where the quantity of description indicates the level of information density. Compared to Sounds and Music, human speech offers a richer array of information, including textual content, emotional expressions, and background noise.

\begin{table}[!h]
    \centering
    \begin{tabular}{cccc}
    \hline
         & Speech & Sound & Music \\
    \hline
    $\Delta$$r$ & 3.78\% & 1.40\% & 0.91\% \\
      $id$  & 3.91 & 1.30 & 2.52 \\
    \hline
    \end{tabular}
    \caption{The analysis of information density focuses on the Speech, Sound, and Music categories. In this analysis, the information density is represented by the number of tokens per second in the description, which is defined as $id$. $\Delta$$r$ represent the $r$ improvements of CoD Prompting compared to Standard Prompting.}
    \label{tab:rst_lalm_2}
\end{table}

We conducted a statistical analysis of the audio in Speech, Sound, and Music categories, calculating the average length of the description generated per second, which is the number of tokens in the description divided by the duration of the audio, defined as $id$. We did not include the Mixed category in our statistics as it is a combination of the other three categories.

The results from Table \ref{tab:rst_lalm_2} indicate that the Speech category can generate an average of nearly 4 tokens of description per second, significantly higher than the Sound and Music categories. Therefore, the use of CoD Prompting in the Speech category yields better results. Although Music has a higher information density than Sound, the improvement when using CoD Prompting in Music and Sound categories is roughly the same, a phenomenon that requires further analysis.

\section{Experiments with CoD in LVLMs}


\subsection{Experimental Setup}

\paragraph{Model}
Our experiments utilized models from the Qwen-VL series, including Qwen2-VL-7B-Instruct\footnote{https://huggingface.co/Qwen/Qwen2-VL-7B-Instruct}, Qwen2.5-VL-7B-Instruct\footnote{https://huggingface.co/Qwen/Qwen2.5-VL-7B-Instruct} and Qwen2.5-VL-72B-Instruct\footnote{https://huggingface.co/Qwen/Qwen2.5-VL-72B-Instruct}.
\paragraph{Evaluation Dataset}
In our experiments, we utilized the MMMU\_Pro\footnote{https://huggingface.co/datasets/MMMU/MMMU\_Pro} dataset. MMMU\_Pro is an enhanced multimodal benchmark designed to rigorously assess the true understanding capabilities of LVLMs. Specifically, we employed the standard validation data within MMMU\_Pro that has been enhanced with 10 options. Additionally, this set can be categorized into three levels of difficulty: Easy, Medium, and Hard. More see Appendix \ref{sec:appendix_case_lvlms}.

\paragraph{Evaluation Method}
As MMMU\_Pro is a multiple-choice dataset, we can directly calculate the accuracy of model predictions. We use both Standard Prompting and CoD Prompting methods to generate the answer options.

\subsection{Results}

\begin{table}[h]
    \centering
    \begin{tabular}{lccc}
    \hline
         & Easy & Medium & Hard \\
    \hline
    \multicolumn{4}{c}{\textit{Qwen2-VL-7B-Instruct}} \\
    \hdashline
    Standard & 40.91\% & 24.91\% & 16.67\% \\
    CoD      & 39.77\% & 22.30\% & \textbf{21.97\%} \\
    \hline
    \multicolumn{4}{c}{\textit{Qwen2.5-VL-7B-Instruct}} \\
    \hdashline
    Standard & 40.91\% & 26.77\% & 16.67\% \\
    CoD      & \textbf{42.61\%} & 26.02\% & \textbf{21.97\%} \\
    \hline
    \multicolumn{4}{c}{\textit{Qwen2.5-VL-72B-Instruct}} \\
    \hdashline
    Standard & 40.91\% & 26.02\% & 18.18\% \\
    CoD      & \textbf{43.18\%} & \textbf{26.02\%} & \textbf{23.48\%} \\
    \hline
    \end{tabular}
    \caption{The evaluation results for LVLMs. We utilized the Qwen2-VL-7B-Instruct and Qwen2.5-VL-7B-Instruct models to conduct tests on the standard 10-option validation set of the MMMU\_Pro dataset. The tests were carried out using both Standard Prompting and CoD Prompting approaches. For simplicity in the results table, we refer to these methods as "Standard" and "CoD" respectively.}
    \label{tab:rst_lvlm_1}
\end{table}




Table \ref{tab:rst_lvlm_1} presents the evaluation results for LVLMs. Among the 7B-sized models, the results indicate that the Qwen2.5-VL-7B-Instruct outperforms the Qwen2-VL-7B-Instruct across all difficulty levels. Surprisingly, CoD Prompting did not demonstrate effectiveness at all levels; instead, it showed significant improvement at the Hard level, increasing the accuracy of both models by 5.3\%. This suggests that CoD Prompting is more beneficial for more complex images or questions. A detailed analysis of the Easy and Medium level cases revealed that due to the huge information density in images, the extensive descriptions generated did not necessarily cover the key points of the questions, potentially leading to a negative impact on responses. Moreover, this negative effect was less pronounced in the latest Qwen2.5-VL-7B-Instruct model.

The Qwen2.5-VL-72B-Instruct outperformed the Qwen2.5-VL-7B-Instruct across all difficulty levels, aligning with the well-known conclusion that larger model sizes lead to stronger performance. Our CoD Prompting method also achieved consistent improvements on the Qwen2.5-VL-72B-Instruct, with a 5.3\% increase at the Hard level. Additionally, it maintained either no decline or a slight improvement at the Easy and Medium levels.

\subsection{Ablation Study}


\begin{mdframed}[backgroundcolor=gray!10]
    \textbf{Question:} \textit{How would the performance be affected if the model could generate higher-quality descriptions?}
\end{mdframed}

The key to our CoD Prompting method lies in generating high-quality and accurate descriptions, which can lead to improved model performance. Knowing that the Qwen2.5-VL-7B-Instruct model outperforms the Qwen2-VL-7B-Instruct, we hypothesize that the descriptions generated by Qwen2.5-VL-7B-Instruct would be superior.

\begin{table}[!h]
    \centering
    \begin{tabular}{lccc}
    \hline
         & Easy & Medium & Hard \\
    \hline
    Standard & 40.91\% & 24.91\% & 16.67\% \\
    CoD      & 39.77\% & 22.30\% & \textbf{21.97\%} \\
    CoD*      & \textbf{42.61\%} & \textbf{25.65\%} & \textbf{21.97\%} \\
    \hline
    \end{tabular}
    \caption{The ablation study for Qwen2-VL-7B-Instruct. We compared two scenarios: in the CoD approach, the description is generated by the model itself; whereas in the CoD* approach, the description is produced by the latest and stronger model Qwen2.5-VL-7B-Instruct.}
    \label{tab:rst_lvlm_2}
\end{table}

We conducted experiments on the Qwen2-VL-7B-Instruct model using descriptions generated by Qwen2.5-VL-7B-Instruct. As shown in Table \ref{tab:rst_lvlm_2}, the results across all difficulty levels demonstrated positive improvements, confirming that higher quality descriptions can yield better outcomes.






\section{Related Work}
We outline some related work and explain how our approach differs from these efforts.

Chain-of-Thought \cite{DBLP:conf/nips/Wei0SBIXCLZ22} (CoT) primarily aims to enhance the reasoning capabilities of text LLMs by breaking down problems into step-by-step solutions. Our method is inspired by CoT, but our CoD is not a variant of CoT; it is a new strategy specifically designed for multi-modal inputs. In theory, CoD and CoT could be combined to bolster the performance of MLLMs.

\citet{DBLP:journals/corr/abs-2311-09193} focuses only on LVLMs and also mentions descriptions. However, their descriptions are related directly to the questions. In contrast, we do focus not on specific questions but explore a more general scenario and verify its effectiveness.
\citet{DBLP:journals/tmlr/0001Z00KS24} also concentrates on LVLMs, but their study emphasizes enhancing reasoning tasks, proposing rationale generation before answer inference. Our work is not limited to reasoning tasks but targets more general scenarios. 
\citet{DBLP:conf/mlcad/VijayaraghavanN24} introduced a method with the same name as ours, but their research direction is about improving the performance of Code LLM.

\section{Conclusion}

This study introduces Chain-of-Description (CoD) Prompting, which significantly enhances the reasoning capabilities of MLLMs. Experiments conducted on models such as Qwen2-Audio, Qwen2-VL, and Qwen2.5-VL demonstrate the effectiveness of CoD Prompting in improving model comprehension and response accuracy for multi-modal inputs, offering a promising direction for future research.


\section{Limitations}

Although our experimental results have been positive, the sheer number of open-source MLLMs and benchmark datasets prevents us from validating each one. Furthermore, based on our motivation, we believe that extensive multi-modal description training during the pre-training phase of MLLMs could yield significant benefits. Regrettably, this type of experiment is particularly resource-intensive, making it infeasible for us to complete swiftly.


\bibliography{custom}

\clearpage
\onecolumn
\appendix
\section{Chain-of-Description Prompting}
\label{sec:appendix1}

\begin{figure*}[!h]
    \centering
    \includegraphics[width=15cm]{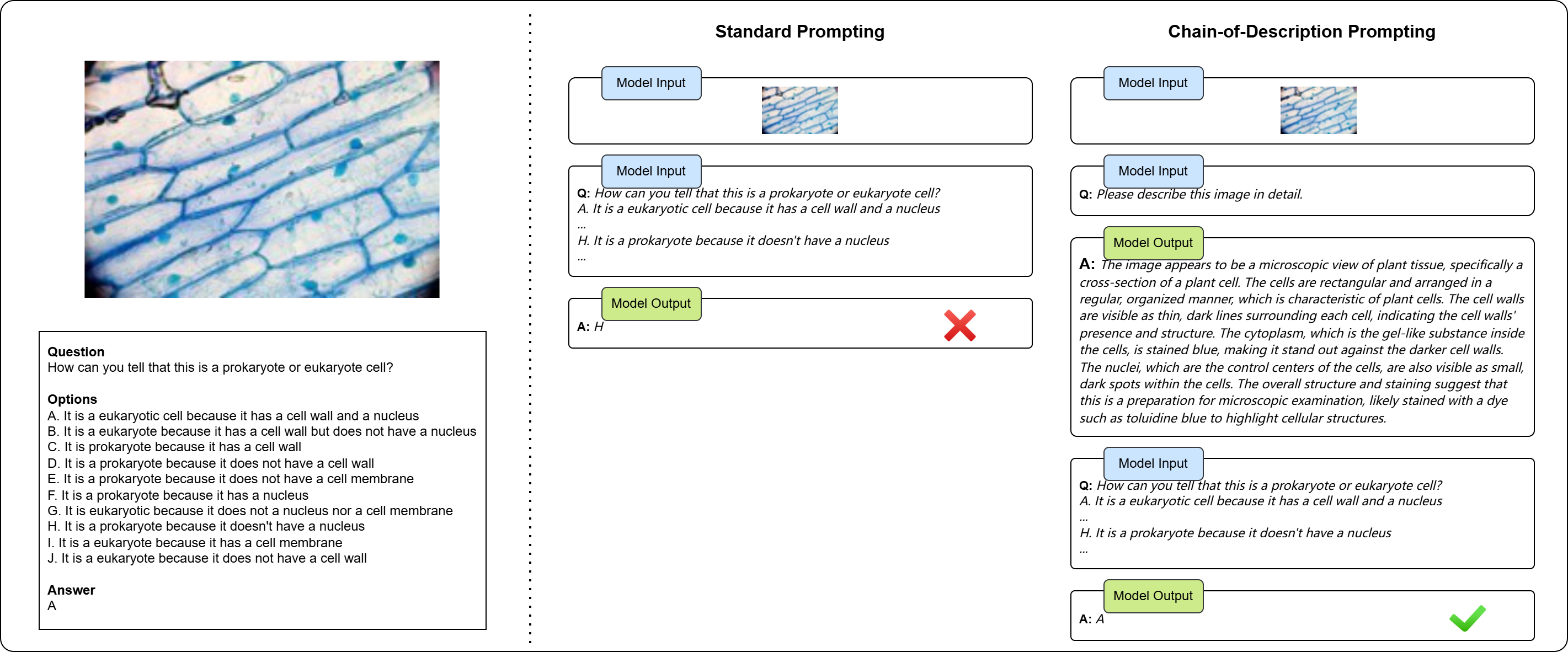}
    \caption{An example of using Standard Prompting and our Chain-of-Description (CoD) Prompting for Large Vision-Language Models (LVLMs).}
    \label{fig:CoD_LVLM}
\end{figure*}

Figure \ref{fig:CoD_LVLM} shows examples of using Standard Prompting and Chain-of-Description Prompting for Large Vision-Language Models.

\clearpage
\section{Statistic and Cases for experiments in LALMs}
\label{sec:appendix_case_lalms}

AIR-Bench-Chat contains 2,200 samples.

\begin{table*}[!h]
\centering
\begin{tabular}{l}
\hline
\textbf{Question} \\
How did the first person react to the second person's situation with restaurants? \\
\hdashline
\textbf{Answer} \\
The first person reacted by saying \colorbox{red}{'wow yeah'}. \\
\hdashline
\textbf{Standard Out} \\
The first person expressed sympathy but also indicated that they do not smoke and are happy about it,\\
suggesting a disconnect from the second person's experience. \\
\hdashline
\textbf{CoD Out} \\
The first person reacted by saying \colorbox{red}{'wow yeah,'} which could indicate surprise or empathy with the \\
second person's situation. \\
\textbf{Cod Description} \\
The audio contains two speakers discussing the topic of smoking and its impact on their rights and \\
choices. The first speaker, who is not a smoker, expresses that they feel it infringes on non-smokers'\\
rights but is happy about the situation since they do not smoke. The second speaker, also not a smoker,\\
lives with smokers and does not believe that smoking should be banned from every place. They \\
mention that they cannot go out to eat at certain restaurants because of smoking policies. The \\ 
conversation ends with both speakers expressing disagreement with the idea of complete smoking \\
bans.\\
\hline
\end{tabular}
\caption{Case 1 for experiments in LALMs}
\label{table:appendix_case1_for_lalms}
\end{table*}

\begin{table*}[!h]
\centering
\begin{tabular}{l}
\hline
\textbf{Question} \\
What kind of weather is mentioned as being present? \\
\hdashline
\textbf{Answer} \\
\colorbox{red}{Rain} is mentioned. \\
\hdashline
\textbf{Standard Out} \\
Colder and damper weather. \\
\hdashline
\textbf{CoD Out} \\
Colder weather and \colorbox{red}{rain} are mentioned. \\
\textbf{Cod Description} \\
The audio contains two people engaging in conversation, discussing the weather and their involvement \\
in a research study. One individual mentions that it is colder where they are and implies that the weather \\
could be related to the study. The other person responds by mentioning that it is damp and rainy where \\
they are, which seems to align with the first person's comment about the cold. Further into the \\
conversation, one of them asks if the other is a college student, to which the other responds affirmatively \\
but clarifies that they have never been linked to a participant before. They both seem surprised and \\
excited about this experience being their first time participating in such a study.\\
\hline
\end{tabular}
\caption{Case 2 for experiments in LALMs}
\label{table:appendix_case2_for_lalms}
\end{table*}

Table \ref{table:appendix_case1_for_lalms} and \ref{table:appendix_case2_for_lalms} are two cases for experiments in LALMs.

\clearpage
\section{Evaluation Prompts of gpt-4o-mini for experiments in LALMs}
\label{sec:appendix2}

\begin{table*}[!h]
\centering
\begin{tabular}{l}
\hline
You are a helpful and precise assistant for checking the quality of the answer. \\
\texttt{[}Detailed Audio Description\texttt{]} \\
XAudioX \\
\texttt{[}Question\texttt{]} \\
XQuestionX \\
\texttt{[}The Start of Assistant 1s Answer\texttt{]} \\
XAssistant1X \\
\texttt{[}The End of Assistant 1s Answer\texttt{]} \\
\texttt{[}The Start of Assistant 2s Answer\texttt{]} \\
XAssistant2X \\
\texttt{[}The End of Assistant 2s Answer\texttt{]} \\
\texttt{[}System\texttt{]} \\
We would like to request your feedback on the performance of two AI assistants in response to the \\
user question and audio description displayed above. AI assistants are provided with detailed audio \\ 
descriptions and questions. \\
Please rate the helpfulness, relevance, accuracy, and comprehensiveness of their responses. Each \\
assistant receives an overall score on a scale of 1 to 10, where a higher score indicates better \\
overall performance. Please output a single line containing only two values indicating the scores for \\
Assistant 1 and 2,
respectively. The two scores are separated by a space. \\
\hline
\end{tabular}
\caption{Evaluation Prompts of gpt-4o-mini for experiments in LALMs}
\label{table:appendix_prompt_for_lalms}
\end{table*}

Table \ref{table:appendix_prompt_for_lalms} shows the evaluation prompts of gpt-4o-mini for experiments in LALMs. "Assistant 1" and "Assistant 2" correspond to the ground truth answer and the model prediction, respectively. To further enhance the fairness of the evaluation, we also swapped the positions of the ground truth answer and the model prediction, then recalculated the final score. 

\clearpage
\section{Evaluation Results for experiments in LALMs}
\label{sec:appendix3}

\begin{table*}[h]
    \centering
    \resizebox{\textwidth}{!}{
    \begin{tabular}{ccccccccccccc}
    \hline
     & \multicolumn{3}{c}{Speech} & \multicolumn{3}{c}{Sound} & \multicolumn{3}{c}{Music} & \multicolumn{3}{c}{Mixed} \\
     & No-S & Swap & Avg & No-S & Swap & Avg & No-S & Swap & Avg & No-S & Swap & Avg \\
    \hline
    \multicolumn{13}{c}{\textit{Standard Prompting}} \\
    \hdashline
    $s_{gt}$ & 8.23 & 8.24 & 8.23 & 7.91 & 8.10 & 8.01 & 8.13 & 7.79 & 7.96 & 8.42 & 8.34 & 8.38 \\
    $s_{gt}$ & 7.63 & 7.39 & 7.51 & 7.03 & 6.82 & 6.93 & 6.95 & 6.79 & 6.87 & 6.50 & 6.48 & 6.49 \\
    $r$(\%) & 92.80 & 89.69 & 91.24 & 88.80 & 84.20 & 86.48 & 85.54 & 87.11 & 86.31 & 77.14 & 77.68 & 77.41 \\
    \hline
    \multicolumn{13}{c}{\textit{CoD Prompting}} \\
    \hdashline
    $s_{gt}$ & 8.11 & 7.98 & 8.04 & 7.81 & 7.73 & 7.77 & 8.22 & 8.02 & 8.12 & 8.24 & 8.00 & 8.12 \\
    $s_{p}$ & 7.81 & 7.47 & 7.64 & 7.01 & 6.64 & 6.83 & 7.17 & 6.99 & 7.08 & 6.40 & 6.36 & 6.38 \\
    $r$(\%) & 96.36 & 93.65 & 95.02 & 89.78 & 85.95 & 87.87 & 87.29 & 87.15 & 87.22 & 77.62 & 79.41 & 78.50 \\
    \hline
    \end{tabular}
    }
    \caption{All evaluation results for experiments in LALMs. "No-S" indicates that "Assistant 1" is the ground truth answer and "Assistant 2" is the model prediction in the evaluation prompts. "Swap" signifies that in the evaluation prompts, "Assistant 2" is the ground truth answer and "Assistant 1" is the model prediction. "Avg" represents the average score between "No-S" and "Swap".}
    \label{tab:appendix_rst_lalm_1}
\end{table*}

Table \ref{tab:appendix_rst_lalm_1} shows the all evaluation results for experiments in LALMs.

\clearpage
\section{Statistic and Cases for experiments in LVLMs}
\label{sec:appendix_case_lvlms}
MMLU\_Pro contains 577 validation samples.

\begin{figure*}[!h]
    \centering
    \includegraphics[width=6cm]{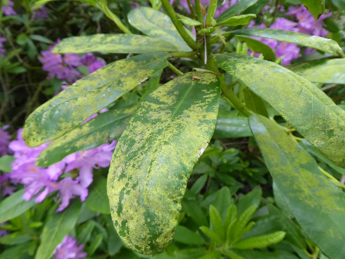}
    \caption{A case image.}
    \label{fig:appendix_case1_img_lvlm}
\end{figure*}

\begin{table*}[!h]
\centering
\begin{tabular}{l}
\hline
\textbf{Question} \\
What is the substance that is developing on these leaves? \\
\hdashline
\textbf{Options} \\
A. Don't know and don't want to guess \ \ B. Powdery mildew \ \ C. Moss \ \ D. Rust \ \ E. Lichen\\
F. Sooty mould \ \ G. Bacterial leaf spot \ \ H. Fungus \ \ I. Downy mildew \ \ J. Algae\\
\hdashline
\textbf{Answer} \\
J. \\
\hdashline
\textbf{Standard Out} \\
B. \\
\hdashline
\textbf{CoD Out} \\
J. \\
\textbf{Cod Description} \\
The image depicts a close-up view of a plant with green leaves that are covered in a yellowish-\\
green substance, possibly \colorbox{red}{algae or a fungal growth}. The leaves are broad and elongated, with \\
visible veins running through them. The plant appears to be healthy overall, with vibrant green \\
foliage and a dense arrangement of leaves. In the background, there are clusters of purple flowers, \\
which are likely part of the same plant or a nearby species. The flowers have multiple petals\\
and are arranged in a dense, bushy manner. The overall scene suggests a natural, outdoor setting,\\
possibly a garden or a forested area. The combination of the green leaves and purple flowers \\
creates a visually appealing contrast. \\
\hline
\end{tabular}
\caption{Cases for experiments in LALMs}
\label{table:appendix_case1_for_lvlms}
\end{table*}

Table \ref{table:appendix_case1_for_lvlms} are cases of Figure \ref{fig:appendix_case1_img_lvlm} for experiments in LALMs.

\end{document}